\documentclass[conference]{IEEEtran}
\IEEEoverridecommandlockouts
\usepackage[colorlinks,
      linkcolor=blue,
      anchorcolor=blue,
      citecolor=blue]{hyperref} % 文献超链接，同时使得跳转部分为蓝色
\usepackage{cite}
\usepackage{amsmath,amssymb,amsfonts}
\usepackage{algorithmic}
\usepackage{algorithm}
\usepackage{graphicx}
\usepackage{subfigure}
\usepackage{textcomp}
\usepackage{xcolor}
\usepackage{siunitx}   
\usepackage{booktabs}
\usepackage{soul}
\usepackage{booktabs}        % For professional horizontal rules
\usepackage{threeparttable}  % For table notes that align with the table width
\usepackage{amssymb}         % For the \checkmark and \times symbols
\usepackage{tikz}
\usetikzlibrary{shapes, arrows.meta, positioning, fit, backgrounds, calc, shadows}
\definecolor{predColor}{RGB}{230, 240, 250} % Light Blue
\definecolor{planColor}{RGB}{255, 250, 230} % Light Yellow
\definecolor{plantColor}{RGB}{235, 250, 235} % Light Green
\definecolor{ctrlColor}{RGB}{240, 240, 240} % Light Gray

\def\BibTeX{{\rm B\kern-.05em{\sc i\kern-.025em b}\kern-.08em
  T\kern-.1667em\lower.7ex\hbox{E}\kern-.125emX}}
\begin{document}
\title{\LARGE \bf SpecFuse: A Spectral-Temporal Fusion Predictive Control Framework for UAV Landing on Oscillating Marine Platforms}
\author{Haichao Liu, Yufeng Hu, Shuang Wang, Kangjun Guo, Jun Ma, \textit{Senior Member, IEEE}, and Jinni Zhou
\thanks{Haichao Liu, Yufeng Hu, Shuang Wang, Kangjun Guo, and Jinni Zhou are with The Hong Kong University of Science and Technology (Guangzhou), Guangzhou 511453, China (e-mail: \{hliu369, yhu444, swang449, kguo886\}@connect.hkust-gz.edu.cn; eejinni@hkust-gz.edu.cn).}
\thanks{Jun Ma is with The Hong Kong University of Science and Technology (Guangzhou), Guangzhou 511453, China, and also with The Hong Kong University of Science and Technology, Hong Kong SAR, China (e-mail: jun.ma@ust.hk).}
% (\textit{Corresponding author: xxx.})}
}

\maketitle

\begin{abstract}
Autonomous landing of Uncrewed Aerial Vehicles (UAVs) on oscillating marine platforms is severely constrained by wave-induced multi-frequency oscillations, wind disturbances, and prediction phase lags in motion prediction. Existing methods either treat platform motion as a general random process or lack explicit modeling of wave spectral characteristics, leading to suboptimal performance under dynamic sea conditions. To address these limitations, we propose SpecFuse: a novel spectral-temporal fusion predictive control framework that integrates frequency-domain wave decomposition with time-domain recursive state estimation for high-precision 6-DoF motion forecasting of Uncrewed Surface Vehicles (USVs). The framework explicitly models dominant wave harmonics to mitigate phase lags, refining predictions in real time via IMU data without relying on complex calibration. Additionally, we design a hierarchical control architecture featuring a sampling-based HPO-RRT* algorithm for dynamic trajectory planning under non-convex constraints and a learning-augmented predictive controller that fuses data-driven disturbance compensation with optimization-based execution. Extensive validations (2,000 simulations + 8 lake experiments) show our approach achieves a 3.2 cm prediction error, 4.46 cm landing deviation, 98.7\%/87.5\% success rates (simulation/real-world), and 82 ms latency on embedded hardware, outperforming state-of-the-art methods by 44\%–48\% in accuracy. Its robustness to wave-wind coupling disturbances supports critical maritime missions such as search and rescue and environmental monitoring. All code, experimental configurations, and datasets will be released as open-source to facilitate reproducibility.
\end{abstract}


\section{Introduction}
\begin{figure}[t]
    \centering
    \subfigure[Operational Schematic Diagram]{
        \includegraphics[width=0.90\linewidth]{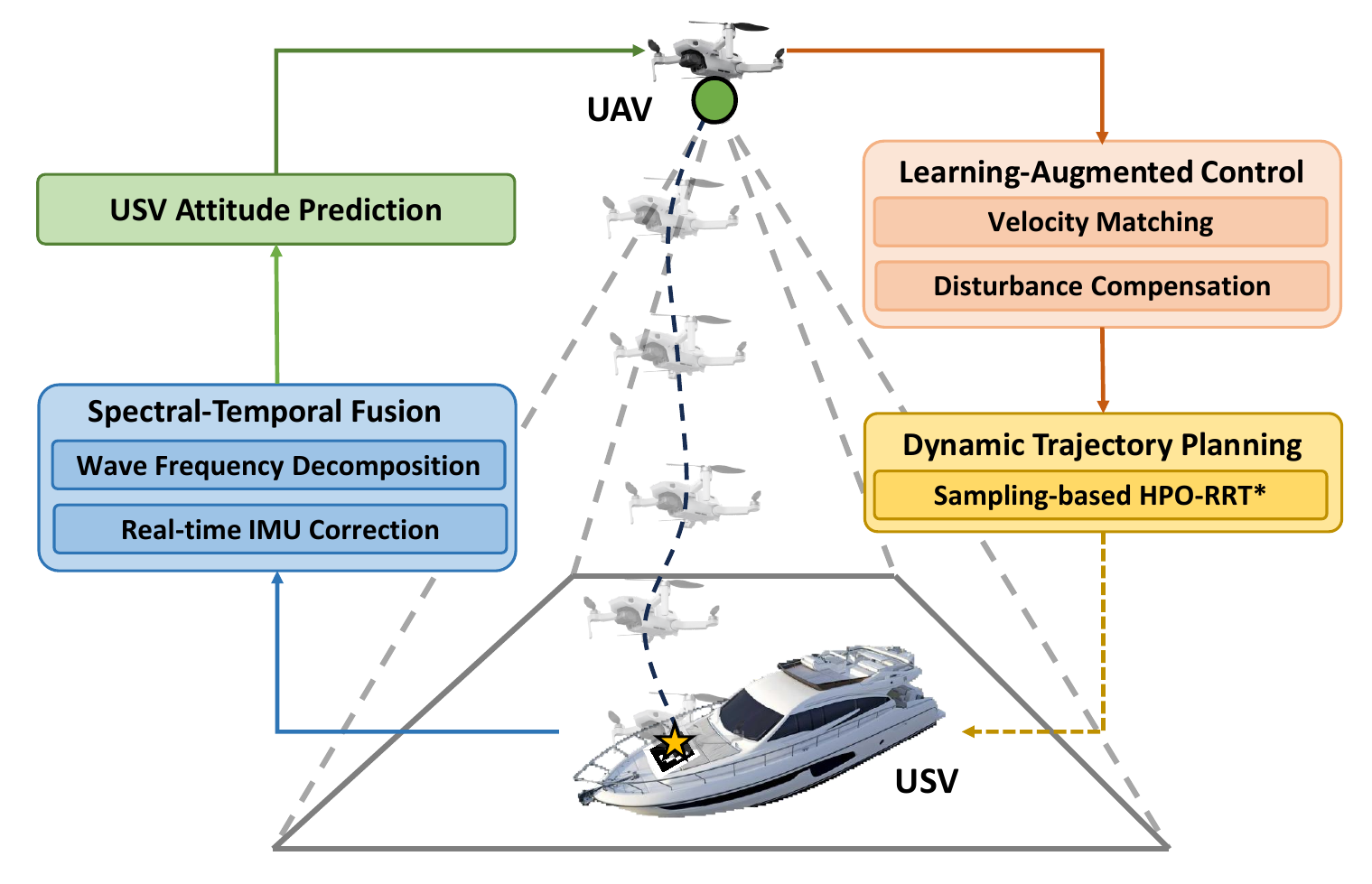}
        \label{fig:teaser_schematic}
    }
    \hfill
    \subfigure[Quantitative Performance]{
        \includegraphics[width=1.0\linewidth]{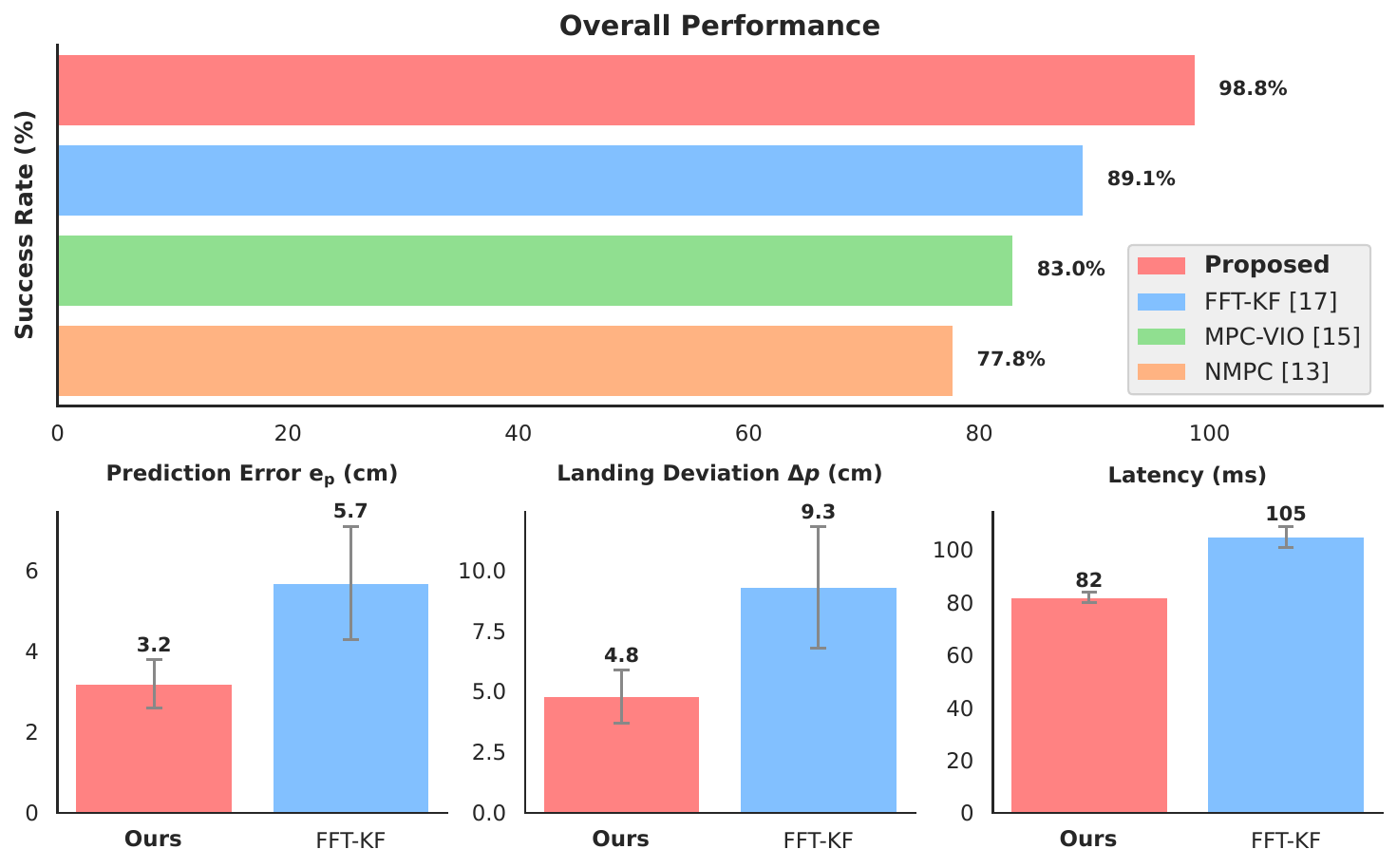}
        \label{fig:teaser_performance}
    }
    \caption{Overview of the UAV-USV maritime rescue operation. The proposed SpecFuse framework (spectral-temporal fusion + learning-augmented control) enables the UAV to predict the USV’s 6-DoF motion (50Hz) and adjust trajectories dynamically, ensuring precise landing ($<$\SI{5}{cm} deviation) for medical resupply missions.}
    \label{fig:teaser_combined}\label{fig:teaser}
\vspace{-0.5cm}
\end{figure}
Autonomous landing of Uncrewed Aerial Vehicles (UAVs) on oscillating marine platforms is a critical capability for maritime missions such as search and rescue, offshore equipment inspection, and emergency material delivery. However, this task is severely challenged by wave, induced multi-frequency oscillations, random wind disturbances, and significant phase lags in motion prediction, key bottlenecks that degrade landing precision and even lead to mission failure or UAV damage in real-world sea conditions~\cite{li2023integrating,8206510,8741624}.
The robotic systems leverage the UAVs' agility and aerial perspective alongside the USVs' extended operational range and payload capacity. Nevertheless, considering the endurance constraints of UAVs, the ability to autonomously land on USVs is critical for prolonged missions. This cooperative task requires precise coordination, yet the dynamic nature of open water environments—characterized by wave-induced oscillations and wind perturbations—introduces significant challenges~\cite{10045543, 9981489,10318328}. To address these difficulties, extensive research has been conducted on dynamic landing control and motion prediction. Traditional methods rely on stochastic estimation or model predictive control (MPC) to handle platform oscillations; however, these approaches often treat motion as a random process and ignore wave spectral characteristics, resulting in unavoidable phase lags. Conversely, while deep learning-based predictors improve generalization, they typically suffer from high computational latency, making them incompatible with embedded UAV hardware.
For instance, Fig.~\ref{fig:teaser} illustrates a schematic of a UAV-USV system for collaborative rescue operations {(e.g., autonomous medical resupply to injured survivors)} in dynamic marine environments, where the UAV must land on {an oscillating marine platform} experiencing rapid pitch and roll oscillations due to disturbances like wave action and strong wind.
UAV landing in dynamic marine environments has been an active research area, with recent advancements focusing on the following three aspects.
(1) Real-time trajectory planning under dynamic disturbances~\cite{8903530,9756345,7470933,10684104} solves the challenge of adapting to the USV's motion.
(2) Sensor fusion for active perching~\cite{10596075,8901098,8421746,10878697,8013706} enhances the UAV's ability to accurately localize and land on moving platforms.
(3) Motion prediction under wave effects~\cite{8484495,9998066} provides foresight into the USV's future states, enabling proactive adjustments to the UAV's trajectory.

While recent advancements like CurviTrack~\cite{gupta2025curvitrack} and MPC-ABCO~\cite{li2025mpc} have demonstrated promising results in controlled environments, they often struggle to adapt to the high uncertainty of open-water operations. Specifically, existing methods cannot simultaneously mitigate phase lags from wave oscillations, ensure real-time performance on embedded hardware, and maintain robustness against wind-wave coupling disturbances. Traditional stochastic estimators~\cite{zhang2025immediate} or deep learning-based predictors~\cite{zheng2025adaptive} frequently treat platform motion as a general random process. This oversight fails to account for the underlying spectral characteristics of ocean waves, resulting in significant tracking errors when the USV encounters multi-frequency oscillations or rapid phase variations~\cite{9998066,10684104}.

To fill this gap, we propose SpecFuse, a novel spectral-temporal fusion predictive control framework for robust UAV landing on oscillating marine platforms. Unlike methods relying solely on reactive control or short-term estimation, SpecFuse leverages a dual-domain architecture: it decomposes motion signals into fundamental spectral components to capture long-term wave periodicities, while simultaneously utilizing a recursive estimator to correct for transient, wind-induced perturbations in the time domain. This integration effectively eliminates phase lag during multi-axis oscillations, enabling precise, computationally efficient trajectory adjustments suitable for embedded systems.

The main contributions are summarized as follows:
\begin{itemize}
  \item {A spectral-temporal motion forecasting framework that integrates frequency-domain wave decomposition with recursive state estimation, explicitly modeling dominant wave harmonics to minimize phase lag on oscillating marine platforms.}
  \item {A hierarchical control architecture utilizing a learning-augmented predictive control scheme, which fuses data-driven disturbance rejection with optimization-based trajectory tracking to ensure precise descent under non-convex constraints.}
  \item {Validation through extensive simulations and field experiments in varying sea states, demonstrating superior landing success rates and prediction robustness on oscillating platforms compared to existing state-of-the-art methods.}
\end{itemize}

\section{Related Works}
Recent advancements in UAV-USV collaborative systems have primarily focused on three critical areas: real-time planning under dynamic disturbances, sensor fusion for active perching, and motion prediction under wave effects. 

\subsection{Real-Time Planning under Dynamic Disturbances}
Real-time trajectory planning under dynamic disturbances is essential for UAV landing on wave-affected platforms~\cite{convens2022safe}. Existing approaches are broadly categorized into heuristic search-based methods, hierarchical optimization strategies, and rolling horizon frameworks. Heuristic search algorithms, such as the local map search by Chen and Gesbert~\cite{8903530}, prioritize efficiency but struggle with highly non-periodic wave motions. Hierarchical optimization strategies~\cite{9756345} decompose complex planning but often lack explicit modeling of wave disturbances. Rolling horizon frameworks~\cite{7470933} enable continuous replanning but may assume quasi-static platform dynamics, failing to accommodate subsecond-scale inclination changes. Wave-induced motion prediction is a key enabler for such planning. Recent work by Liu et al.~\cite{10684104} combined spectral analysis with adaptive filtering, achieving high accuracy and low latency. While this provides a foundation, it does not fully address the integration of spectral characteristics into a multi-resolution planning architecture.

\subsection{Sensor Fusion for Active Perching}
In the domain of sensor fusion for active perching, various strategies have been explored. Bio-inspired control~\cite{10596075} and Nonlinear optimization~\cite{8901098,liu2025udmc,li2025mpc} have shown success in robotics but face reduced effectiveness in outdoor crosswinds or when encountering multiple obstacles. Advances in localization, such as VINS-Mono~\cite{8421746} and lightweight depth estimation networks~\cite{10878697}, must trade off millimeter-level accuracy with high computational latency or robustness to illumination changes. Similarly, moving platform tracking systems~\cite{8013706} have demonstrated successful landings but can exhibit significant overshoots and sensor fusion latency during sudden maneuvers.

\subsection{Attitude Prediction under Wave Effects}
Accurate prediction of wave or uncrewed vessel attitude is crucial for developing control strategies. While Fourier transforms have been applied to predict short-term wave behavior~\cite{8484495}, these predictions often show phase misalignment over extended periods. {Optimization-based} methods have been proposed for landing on ships in rough seas~\cite{9998066,gupta2025curvitrack}, achieving strong short-term correlation, but prediction accuracy degrades progressively in forecasts beyond one second. Notably, many existing models employing spectral methods for state prediction face challenges in achieving sufficient alignment of phase and amplitude, which is a critical factor limiting prediction accuracy.
While significant progress has been made, challenges remain in accurately predicting multi-axis oscillations and rapid phase variations caused by wind-wave coupling.
Our work builds on these advancements by integrating UAV landing path planning with {oscillating marine platform} attitude prediction using a {spectral-temporal fusion} approach, enabling real-time trajectory adjustments based on accurate motion forecasts.
\section{Methodology}
\begin{figure}[t]
\centering
\resizebox{\columnwidth}{!}{%
\begin{tikzpicture}[
  node distance=1.0cm and 0.4cm, 
  % Box Styles
  block/.style={rectangle, draw=black!70, fill=white, text width=1.8cm, minimum height=1.0cm, align=center, rounded corners=1pt, font=\footnotesize\sffamily},
  % Module container style
  module/.style={draw, dashed, inner sep=8pt, rounded corners=4pt, font=\bfseries\small\sffamily},
  % Line styles
  line/.style={-Latex, draw=black!80, semithick},
  % Text Label style
  label text/.style={font=\scriptsize\itshape, text=black!80, inner sep=2pt, align=center} 
]

  % --- 1. Top Stream ---
  \node (fft) [block] {Spectral\\ Decomp.};
  
  \node (kf) [block, right=0.8cm of fft] {Recursive\\ Estimator};

  \node (mpc) [block, right=1.0cm of kf, text width=2.2cm, double] {Learning-Augmented\\ Controller};
  
  \node (uav) [block, right=1.0cm of mpc] {Actuators\\\& Dynamics};
  \node (sensors) [block, below=0.5cm of uav] {Onboard\\Sensors};

  % --- 2. Bottom Stream ---
  \node (constraints) [block, below=1.8cm of kf] {Constraints\\Handling};
  \node (hpo) [block, left=of constraints] {HPO-RRT*\\Optimization};
  \node (formulation) [block, left=of hpo] {Problem\\Formulation};
  \node (state_est) [block, right=of constraints, xshift=1.115cm] {State\\Estimator};

  % --- 3. Connections & Labels ---
  \node (usv_input) [left=0.4cm of fft, font=\footnotesize\sffamily, align=right] {Platform Data};
  \draw[line] (usv_input) -- (fft);

  \draw[line] (fft) -- node[label text, above] {Freq.\\Comp.} (kf);
  \draw[line] (kf) -- node[label text, pos=0.6, above] {$\hat{\mathbf{x}}_\text{USV}$} (mpc);
  
  \draw[line] (mpc) -- node[label text, pos=0.35, above] {$u_\text{cmd}$} (uav);
  \draw[line] (uav) -- (sensors);

  \draw[line] (sensors) |- (state_est);
  \draw[line] (state_est.north) -- ++(0,0.5) -| node[label text, pos=0.85, right] {$\mathbf{r}_\text{UAV}$} (mpc.300); 

  \draw[line] (formulation) -- (hpo);
  \draw[line] (hpo) -- (constraints);

  % Inter-Module Connections
  \draw[line] (kf.south) -- ++(0,-0.8) -| node[label text, above, pos=0.2] {Motion Forecast} (formulation.north);
  
  \draw[line] (constraints.north) -- ++(0,0.5) -| node[label text, left, pos=0.85] {$\mathbf{r}_\text{ref}$ } (mpc.240);
  
  \draw[line] (state_est.south) -- ++(0,-0.5) -| node[label text, below, pos=0.2] {$\mathbf{r}_\text{UAV}$} (formulation.south);

  % --- 4. Background Modules ---
  \begin{scope}[on background layer]
    \definecolor{predColor}{RGB}{235, 245, 255}
    \definecolor{planColor}{RGB}{255, 252, 235}
    \definecolor{plantColor}{RGB}{240, 255, 240}

    \node [module, fill=predColor, fit=(fft) (kf) (usv_input), label={[anchor=south west, inner sep=2pt]north west:Platform Motion Prediction}] {};
    
    \node [module, fill=planColor, fit=(formulation) (hpo) (constraints), 
           label={[anchor=south west, xshift=1.4cm, inner sep=2pt]north west:Trajectory Planner}] {};
    
    \node [module, fill=plantColor, fit=(uav) (sensors), label={[anchor=south east, inner sep=2pt]north east:UAV Plant}] {};
  \end{scope}

\end{tikzpicture}
}
\vspace{-0.5cm}
\caption{Proposed control architecture. The Motion Prediction module forecasts the platform's state, while the Learning-Augmented controller utilizes multi-source feedback to command the UAV actuators.}
\label{pipeline}
\vspace{-0.5cm}
\end{figure}
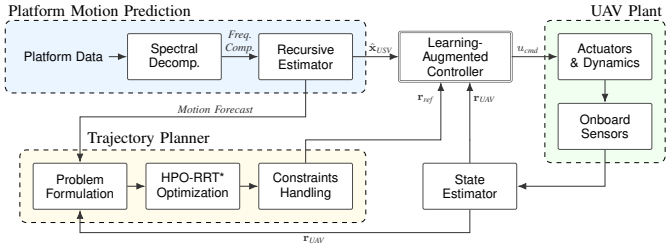

The overall architecture of our aerial system in dynamic marine environments is illustrated in Fig.~\ref{pipeline}. As illustrated in Algorithm~\ref{alg:workflow}, the system operates through a closed-loop architecture where IMU/GNSS data from the {oscillating platform} is continuously fed into a motion prediction module (Section~\ref{subsec:motionPredictionUSV}). This module generates 6-DoF motion estimates at a frequency of 50 Hz, which are then used by the trajectory planner (Section~\ref{subsec:trajectoryPlanningUAV}) to dynamically adjust the UAV’s flight path. The execution controller (Section~\ref{subsec:executionControllerUAV}) implements the optimized trajectory using {learning-based regression and receding horizon optimization} algorithms.

\subsection{Motion Prediction for {Oscillating Marine Platforms}}
\label{subsec:motionPredictionUSV}

This module estimates the platform’s 6-DoF motion in real time by fusing frequency-domain spectral decomposition with adaptive recursive estimation. This approach leverages the quasi-periodicity of hydrodynamic wave interactions, where dominant energy frequencies remain stationary during the short landing window. While a formal topological proof is omitted, the estimator’s convergence is practically secured by the bounded nature of wave amplitudes and the contraction properties of the update law. This enables accurate 50 Hz motion forecasting, which is critical for compensating wave-induced oscillations.+4The prediction process follows three core steps:
\textbf{Spectral Decomposition} extracts amplitude $A$, phase $\varphi$, and frequency $f$ by decomposing wave signals into harmonic components.
\textbf{Recursive State Estimation} utilizes previous spectral results as a prior model and refines them using real-time IMU data to obtain filtered wave parameters.
\textbf{Conditioned Prediction} leverages the refined parameters $A, f, \varphi$ to forecast future deck attitudes over the planning horizon.

\subsubsection{{Spectral Decomposition of Wave Motion}}
Since the 6-DoF attitude of the {oscillating platform} can be decomposed into independent components in each direction~\cite{10874309}, we take the signal of the component roll around the $x$-axis as an example for prediction. First, to determine the periodicity of the signal, we take a segment $\Delta t$ ahead of the current signal time $t$ as a measure of periodicity, i.e., $\mathcal{T}=[t-\Delta t, t]$, and traverse the previous signals to judge which moment $T\in\mathcal{T}$ of the signal has the strongest similarity to this segment. The Euclidean distance of the time series is used to determine this similarity~\cite{6950920}.

Consider a time series $T_1$ containing $K$ sampling points prior to the prediction start time point $T$. We find a segment of time series $T_2$ that also contains $K$ sampling points and is most similar in shape to time series $T_1$. The difference between the two curves is measured by the Euclidean distance:
\begin{equation}
Eu_\text{dist}(T_{1i}, T_{2i}) = |T_{1i}(y) - T_{2i}(y)|,
\end{equation}
where $T_{\dot i}$ denotes the $i$-th sampling point in time series $T$. The average Euclidean distance over $K$ sampling points is calculated as:
\begin{equation}
\overline{Eu_\text{dist}} = \frac{1}{k} \sum_{i=1}^{K} Eu_\text{dist}(T_{1i}, T_{2i}).
\end{equation}
The variance of the $Eu_\text{dist}$ is recorded as $\sigma_K(Eu_\text{dist})$:
\begin{equation}
\sigma_K(Eu_\text{dist}) = \frac{1}{k} \sum_{i=1}^{K} \left( Eu_\text{dist}(T_{1i}, T_{2i}) - \overline{Eu_\text{dist}} \right)^2.
\end{equation}
A smaller value of $\sigma_K\big(Eu_{\text{dist}}\big)$ indicates a greater degree of similarity between $T_1$ and $T_2$. The {spectral decomposition} is then expressed as:
\begin{equation}
X(k) = \sum_{n=0}^{N-1} x(n) e^{-j2\pi kn/N}, \quad k = 0, 1, \ldots, N-1,
\end{equation}
where \( X(k) \) is the discrete transform of the signal \( x(n) \), and \( N \) is the number of samples.
The transform is computed for the time-series signal within the interval ${[T, t]}$. To mitigate the influence of noise, frequency components with an amplitude ${A}$ less than ${0.02 A_{\max}}$ are omitted. This filtered result is then used to construct the prediction model for the signal after time $t$. The underlying premise of this method is the assumed {stationarity or periodicity} of the signal, which implies that the filtered spectrum of the segment ${[T, t]}$ is a faithful representation of the signal's spectral content for ${t' > t}$.

\subsubsection{{Recursive State Estimation}}
We estimate the state using the roll direction signal~\cite{9842272}.  This signal is mathematically modeled as a superposition of independent sinusoidal components (sine waves), where the signal is defined as: $A_{i}\sin(2\pi f_{i}(t) + \varphi_{i}), i=1,2,3...,N$. In this model, each component $i$ is considered an independent harmonic oscillator. The instantaneous state of each oscillator, $v_i$, which comprises its position and velocity, and its dynamics (time derivative), $v'_i$, are defined as follows:
\begin{equation}
v_{i} = \begin{bmatrix}
A_{i}\,\sin(\Phi) \\
2\pi f_{i} A_{i} \cos(\Phi)
\end{bmatrix},
\end{equation}
\begin{equation}
v'_{i} = \underbrace{\begin{bmatrix} 0 & 1 \\ -(2\pi f_{i})^2 & 0 \end{bmatrix}}_{\mathbf{B}(i)} v_{i},
\end{equation}
where $A_{i}$ and $f_{i}$ are the parameters derived from the {spectral analysis}. The observation information is the IMU attitude and angular velocity.

This forms the basis of the {adaptive recursive estimation} process. The full state vector $\hat{x}$ (as seen in Eq.~\ref{eq:kf_predict}) is constructed by augmenting the individual state vectors $v_i$ for all $N$ components, and the full state transition model $A$ is a block-diagonal matrix composed of the individual $\mathbf{B}(i)$ matrices. This design assumes the sinusoidal components are independent and allows the {estimator} to refine the amplitude and phase of each component in real-time.

The core estimation process involves the steps of prediction and measurement update. 
Note taht the measurement update incorporates the IMU observations to refine the state estimate.
The state prediction can be expressed as:
\begin{equation}
\label{eq:kf_predict}
\hat{x}_{k|k-1} = A \hat{x}_{k-1|k-1} + B u_{k-1}.
\end{equation}
In addition, covariance prediction is given by:
\begin{equation}
P_{k|k-1} = A P_{k-1|k-1} A^T + Q.
\end{equation}
The measurement update at step $k$ is defined as:
\begin{equation}
  \label{eq:kf_update}
\left\{
\begin{aligned}
y_k &= z_k - H \hat{x}_{k|k-1} \\
S_k &= H P_{k|k-1} H^T + R \\
K_k &= P_{k|k-1} H^T S_k^{-1}\\
\hat{x}_{k|k} &= \hat{x}_{k|k-1} + K_k y_k \\
P_{k|k} &= (I - K_k H) P_{k|k-1},
\end{aligned}
\right.
\end{equation}
where $\hat{x}$ represents the state estimate; $P$ is the estimate covariance matrix; $A$ and $B$ are the state transition and control-input models, respectively, with $u$ being the control vector; $Q$ is the process noise covariance; $z$ is the measurement vector, which yields the measurement residual $y$; $H$ is the observation model; $R$ is the measurement noise covariance; and $K$ is the computed gain.

\subsubsection{Condition Prediction}
Given the last observation time $t_{\text{obs}}$, the execution of the {state estimator}, as defined by (\ref{eq:kf_predict}) to (\ref{eq:kf_update}), provides the current state parameters $A_{j,i}(t_{\text{obs}})$ and $\varphi_{j,i}(t_{\text{obs}})$. These resulting values are then used, in conjunction with {spectral} characteristics, to model and describe the subsequent wave propagation state:
\begin{equation}
\Phi_{i}(t_{obs}) = \arctan \left( \frac{2\pi f_{i} [\mathbf{v}_{i}]^{1,1}}{[\mathbf{v}_{i}]^{2,1}} \right),
\end{equation}
and
\begin{equation}
A_{i}(t_{obs}) = \frac{[\mathbf{v}_{i}]^{1,1}}{\sin \left( \Phi_{i}(t_{obs}) \right)}.
\end{equation}
The calculated parameters can then be used to make predictions for moments $t < t_{obs}$:
\begin{equation}
\begin{split}
b(t) = & \sum_{i=1}^{N} A_{i}(t_{obs})\,\sin[2\pi f_{i}(t - t_{obs}) + \varphi_{i}(t_{obs})].
\end{split}
\end{equation}
The application of this technique facilitates the prediction of the {oscillating marine platform's} future movement over a brief time horizon.

\begin{algorithm}[t]
\caption{Predictive-Coordination of Robust Landing Control on {Oscillating Marine Platforms}}
\label{alg:workflow}
\begin{algorithmic}[1]
\REQUIRE UAV state $s_0 = [x_0, y_0, z_0, \dot{x}_0, \dot{y}_0, \dot{z}_0, \phi_0, \theta_0, \psi_0]$, \\
\quad Platform motion data $D_{\text{USV}}$, environmental parameters $E$

\STATE \textbf{Initialization:}
\STATE \quad $A_i, f_i, \varphi_i \gets \text{SpectrumDecomp}(D_{\text{USV}})$.
\STATE \quad $d_{\text{safe}} \geq 0.5\, \mathrm{m}$, $\phi, \theta \in [-45^\circ, 45^\circ]$.

\STATE \textbf{Prediction Phase:}
\STATE \quad $D_{\text{USV}} \gets \text{IMU/GNSS}$.
\STATE \quad $D_{\text{USV}} = \sum_{i=1}^N A_i\,\sin(2\pi f_i t + \varphi_i)$.
\STATE \quad $\hat{x}_t = \text{RecursiveEst}(A_i, f_i, \varphi_i, D_{\text{USV}})$.
\STATE \quad $\hat{x}_t = [\hat{x}, \hat{y}, \hat{z}, \hat{\phi}, \hat{\theta}, \hat{\psi}]$.

\STATE \textbf{Trajectory Planning Phase:}
\STATE \quad $\text{Traj} \gets \text{HPO-RRT*}(s_0, \hat{x}_t)$.
\STATE \quad $\text{Traj} = \arg\min_{u} \int_{t_0}^{t_f} L(s, u) \, dt$.
\STATE \quad $\text{Traj} \gets \text{Traj} + \Delta \text{Traj}(\hat{x}_t)$.

\STATE \textbf{Execution Phase:}
\STATE \quad $\min_{w, \xi} \frac{1}{2} \|w\|^2 + C \sum_{i=1}^n (\xi_i + \xi_i^*)$.
\STATE \quad $u^* = \arg\min_{u} \int_{t_0}^{t_f} L(s, u) \, dt$.
\STATE \quad $P_{\text{conf}} \geq 80\%$, $V_{\text{batt}} \geq 21.0\, \mathrm{V}$.

\STATE \textbf{Fail-Safe Mechanism:}
\IF{$P_{\text{conf}} < 80\%$ \OR $V_{\text{batt}} < 21.0\, \mathrm{V}$}
  \STATE Trigger emergency hover: $\dot{x} = \dot{y} = \dot{z} = 0$.
\ENDIF

\STATE \textbf{Landing:}
\STATE \quad $\dot{x}_{\text{UAV}} = \dot{x}_{\text{USV}}$, $\dot{y}_{\text{UAV}} = \dot{y}_{\text{USV}}$.
\STATE \quad $z_{\text{UAV}} \to z_{\text{USV}}$.

\RETURN Successful landing.
\end{algorithmic}
\end{algorithm}

\subsection{SE(3) Trajectory Planning for the UAV}\label{subsec:trajectoryPlanningUAV}
The trajectory planner employs the HPO-RRT* algorithm~\cite{9512957} to generate SE(3) trajectories optimized for dynamic environments. The planner dynamically adjusts trajectories based on real-time motion predictions, ensuring collision-free and time-efficient paths. An SE(3) trajectory can be represented as a function \( \tau(t) \) that maps time \( t \) to a configuration in the special Euclidean group:
\begin{equation}
\tau(t) = \begin{bmatrix} \mathbf{R}(t) & \mathbf{p}(t) \\ 0 & 1 \end{bmatrix},
\end{equation}
\noindent where \( \mathbf{R}(t) \in SO(3) \) is the rotation matrix and \( \mathbf{p}(t) \in \mathbb{R}^3 \) is the translation vector.

For SE(3) motion planning, we implement a time-optimal point-mass model trajectory planning approach that considers the UAV's rotational dynamics and physical constraints~\cite{10239764}. The goal is to generate a trajectory that is both time-efficient and safe. The point-mass model is defined by the following state variables:
\begin{equation}\,\mathbf s = [x, y, z, \dot{x}, \dot{y}, \dot{z}, \phi, \theta, \psi]^T, \end{equation}
where $x,y,z$ are the position coordinates, $\dot{x}, \dot{y}, \dot{z}$ are the velocity components, and $\phi, \theta, \psi$ are the roll, pitch, and yaw angles, respectively.
The trajectory planning problem is formulated as an optimization problem:
\begin{equation}
\begin{aligned}
& \min_{\mathbf u} \int_{t_0}^{t_f} L(\mathbf s, \mathbf u) \, dt \\
& \text{s.t.} \quad \dot{\mathbf s} = f(\mathbf s, \mathbf u), \\
& \phantom{\text{s.t.}} \quad g(\mathbf s, \mathbf u) \leq 0,
\end{aligned}
\end{equation}
where $\mathbf u$ is the control input, $L(\mathbf s,\mathbf u)$ is the cost function, $f(\mathbf s,\mathbf u)$ is the system dynamics, and $g(\mathbf s,\mathbf u)$ represents the physical and safety constraints. To solve this optimization problem, we use a sequential quadratic programming (SQP) approach that iteratively refines the trajectory while ensuring feasibility.

The HPO-RRT* algorithm extends the Probabilistic Roadmap (PRM) method by incorporating heuristic sampling and optimization. It features a hierarchical architecture with three core components: \textbf{Threat prediction} utilizes the {learning-based} model to predict platform and obstacle motion, updating the dynamic obstacle map in real time. \textbf{Dynamic path planner} employs the Improved Time-Based RRT* to generate initial paths that minimize time while avoiding collisions with predicted obstacles. \textbf{Trajectory optimizer} refines the initial path using a Low-Cost Path Optimizer that performs both coarse and fine optimizations to ensure smoothness and feasibility. The three components work synergistically to produce optimal trajectories in dynamic environments:
The cost function for the Improved Time-Based RRT* is defined as:
\begin{equation}
C_{\text{total}} = \alpha t + \beta d + \gamma m,
\end{equation}
where $t$ is path time, $d$ denotes path length, $m$ represents safety margin, and \( \alpha, \beta, \gamma \) are their respective weighting parameters. This formulation enables the generation of a time-efficient and safe path for the UAV.

The cost function for the overall trajectory optimization can also be formulated as:
\begin{equation}
J(\tau) = \int_{t_0}^{t_f} \left( \mathbf{u}^T \mathbf{Q} \mathbf{u} + \mathbf{v}^T \mathbf{R} \mathbf{v} \right) dt,
\end{equation}
subject to key operational constraints. These include thrust limits of \( u_{\text{max}} = 15 \, \text{N} \) per motor, safety margins requiring a minimum distance of \( \geq 0.5 \, \text{m} \) from the platform deck, and physical bounds restricting pitch and roll angles to \( \pm 45^\circ \). These constraints are formulated as:
\begin{equation}
\left\{
\begin{aligned}
& \| \mathbf{p}(t) - \mathbf{p}_{\text{USV}}(t) \| \geq 0.5 \, \text{m} \\
& |\phi(t)| \leq 45^\circ \\
& |\theta(t)| \leq 45^\circ \\
& \| \mathbf{u}(t) \| \leq 15 \, \text{N}
\end{aligned}
\right.
\end{equation}
where \( \mathbf{u} \) is the control input, \( \mathbf{v} \) is the velocity, \( \mathbf{Q} \) and \( \mathbf{R} \) are positive definite weighting matrices, \( \mathbf{p}_{\text{USV}} \) is the position of the platform, and \( \phi \) and \( \theta \) are the pitch and roll angles.

\subsection{Execution Controller for the UAV}
\label{subsec:executionControllerUAV}
This module implements a hybrid control strategy combining {learning-based} velocity matching and {receding horizon control}. The {learning model} minimizes the tangential relative speed between the UAV and {oscillating platform}, while the {receding horizon optimizer} handles the UAV's attitude and position during the final descent.

\subsubsection{{Learning-Based Velocity Matching}}
To minimize the tangential relative speed between the UAV and the platform, we employ a {data-driven regression} model with mixed kernel functions~\cite{10321664}. This method enables precise prediction of the platform's motion, which is critical for adjusting the UAV's approach velocity to match the movement during the final approach phase. The model is formulated as follows:
\begin{equation}
\begin{aligned}
& \min_{w, b, \xi, \xi^*} \quad \frac{1}{2} \|w\|^2 + C \sum_{i=1}^n (\xi_i + \xi_i^*), \\
& \text{s.t.} \quad y_i - w^T \phi(x_i) - b \leq \epsilon + \xi_i, \\
& \phantom{\text{s.t.}} \quad w^T \phi(x_i) + b - y_i \leq \epsilon + \xi_i^*, \\
& \phantom{\text{s.t.}} \quad \xi_i, \xi_i^* \geq 0, \quad \text{for } i = 1, \dots, n,
\end{aligned}
\end{equation}
where $w$ is the weight vector, $\xi$ and $\xi^*$ are slack variables, $C$ is the regularization parameter, $\epsilon$ is the insensitive loss parameter, $\phi(x)$ is the kernel function, and $y_i$ is the target output.
To improve prediction accuracy, we use a mixed kernel function that combines the radial basis function (RBF) kernel and the polynomial kernel:
\begin{equation} \phi(x) = \alpha \cdot \text{RBF}(x) + (1 - \alpha) \cdot \text{Polynomial}(x), \end{equation}
where $\alpha$ is a weighting parameter that balances the contributions of the two kernels. This hybrid kernel allows the {regressor} to capture both the nonlinear dynamics of the platform motion and the local geometric features of the trajectory. The predicted motion is used to adjust the UAV's velocity vector, ensuring that the tangential relative speed is minimized during the final approaching period.

\subsubsection{{Receding Horizon Descent Control}}
The {receding horizon} formulation involves minimizing the cost function:
\begin{equation}
\begin{split}
J_{\mathbf{r,u}} = \sum_{k=0}^{N_p-1} \Big( & \|\mathbf{r}(t + k T_s) - \mathbf{r}_{\text{ref}}(t + k T_s)\|^2_Q \\
& + \|\Delta \mathbf{u}(t + k T_s)\|^2_R \Big) \\
& + \|\mathbf{r}(t + N_p T_s) - \mathbf{r}_{\text{ref}}(t + N_p T_s)\|^2_P.
\end{split}
\end{equation}
Therefore, the corresponding optimization problem can be expressed as:
\begin{equation}
\begin{aligned}
\min_{\mathbf{u}} \quad & J_{\mathbf{r,u}}, \\
\text{s.t.} \quad & \mathbf{r}(t + (k+1) T_s) = f(\mathbf{r}(t + k T_s), \mathbf{u}(t + k T_s)), \\
& \mathbf{u}_{\text{min}} \leq \mathbf{u}(t + k T_s) \leq \mathbf{u}_{\text{max}}, \\
& \Delta \mathbf{u}_{\text{min}} \leq \Delta \mathbf{u}(t + k T_s) \leq \Delta \mathbf{u}_{\text{max}}, \\
& \quad \text{for } k = 0, \dots, N_p-1.
\end{aligned}
\end{equation}
where \( \mathbf{r} \) is the state vector, \( \mathbf{u} \) is the control input, \( T_s \) is the sampling time, \( N_p \) is the prediction horizon, \( \mathbf{r}_{\text{ref}} \) is the reference trajectory, \( f \) is the system dynamics function, and \( \Delta \mathbf{u} \) is the control input increment.
\begin{figure}[t]
\centering
\resizebox{2.7in}{1.8in}{\includegraphics{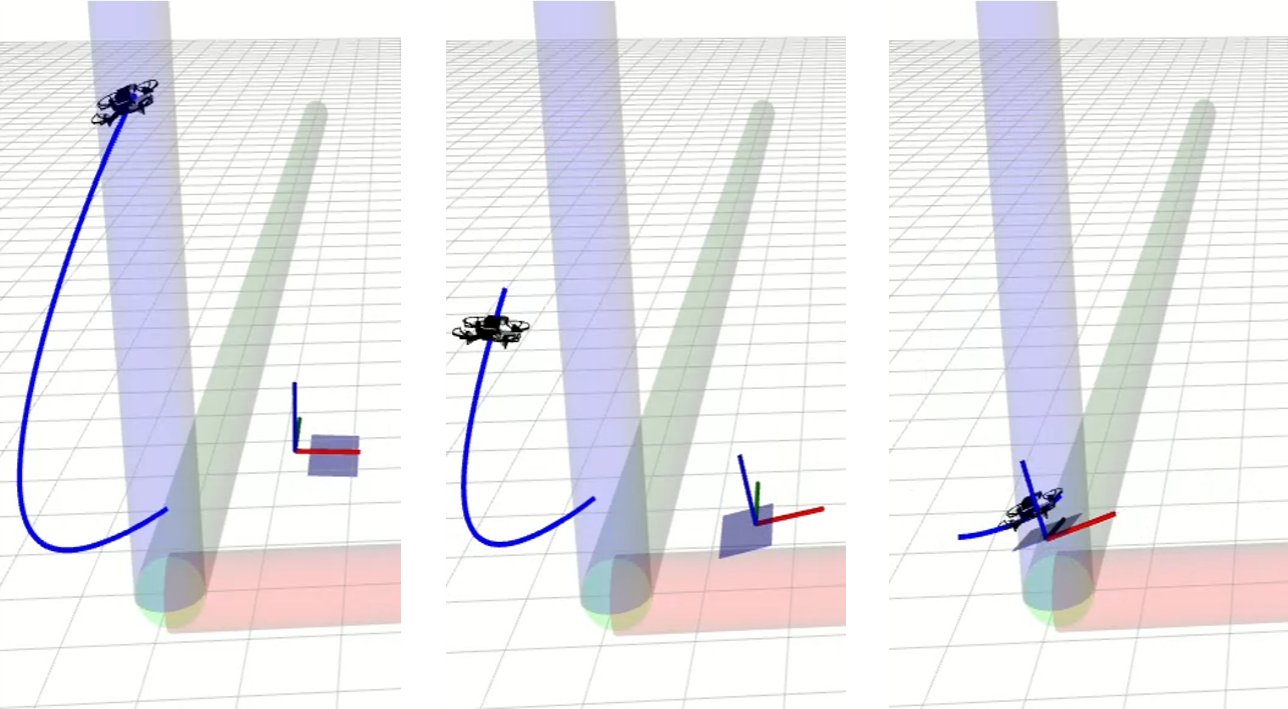}}
\caption{Numerical validation in simulation of adaptive trajectory planning for UAV/USV cooperative landing under high-fidelity marine disturbances. The blue line represents the UAV’s planned trajectory, while the square platform with coordinate axes indicates the pose of the {oscillating marine platform}.
}
\label{fig:replanning}
\vspace{-0.3cm}
\end{figure}
\section{Experimental Results and Analysis}
\label{sec:experiments}
\begin{table}[t]
\caption{Simulation Performance Comparison with Challenging Marine Disturbances}
\label{tab:benchmark}
\centering
\resizebox{\columnwidth}{!}{%
\begin{tabular}{lcccc}
\toprule
\textbf{Method} & $\boldsymbol{e_p}$ (\si{cm}) $\downarrow$ & $\boldsymbol{\Delta p}$ (\si{cm}) $\downarrow$ & \textbf{SR} $\uparrow$ & \textbf{Latency} (\si{ms}) $\downarrow$ \\
\midrule
FFT-KF~\cite{8484495} & 5.7 $\pm$ 1.4 & 9.3 $\pm$ 2.5 & 89.1\% & 105 $\pm$ 4 \\
MPC-VIO~\cite{10878697} & 6.4 $\pm$ 1.9 & 11.8 $\pm$ 3.3 & 83.0\% & 118 $\pm$ 3 \\
NMPC~\cite{8901098} & 8.9 $\pm$ 2.8 & 14.1 $\pm$ 4.1 & 77.8\% & 142 $\pm$ 5 \\
% \midrule
Proposed & \textbf{3.2 $\pm$ 0.6} & \textbf{4.8 $\pm$ 1.1} & \textbf{98.8\%} & \textbf{82 $\pm$ 2} \\
\bottomrule
\end{tabular}
}
\vspace{-0.3cm}
\end{table}

To validate the proposed predictive-coordination approach for the landing control of UAVs onto {oscillating marine platforms} operating in dynamic conditions, a comprehensive evaluation encompassing both simulations and real-world experiments was conducted under challenging scenarios.
The evaluation strategy comprises distinct components: \textbf{High-fidelity simulation studies} to establish performance baselines and facilitate comparative analysis against existing landing control methodologies; and \textbf{Challenging outdoor lake experiments} serving to test the system's robustness and operational efficacy within unstructured, real-world maritime environments.

\subsection{Simulation Results}
\subsubsection{Experimental Setup and Baselines}
The simulation environment was developed within Gazebo. Wave disturbances were generated utilizing the Joint North Sea Wave Project (JONSWAP) spectrum \cite{rueda2020selection}. The specific parameters were configured to represent challenging maritime conditions, defined by a significant wave height of $H_s = 1.8 \, \text{m}$ and a peak period of $T_p = 6.5 \, \text{s}$. Wind disturbances were modeled using the Dryden turbulence model~\cite{ko2025wind}, which provided a realistic representation of atmospheric conditions. The model was parameterized with a mean wind velocity of $v_{\text{mean}} = 8 \, \text{m/s}$ and a turbulence intensity (standard deviation) of $\sigma = 2.5 \, \text{m/s}$.

We compared our proposed method against three baselines: the frequency-domain \textbf{FFT-KF}~\cite{8484495}, a nonlinear model predictive control (\textbf{NMPC})~\cite{8901098}, and a visual-inertial odometry-enhanced \textbf{MPC-VIO}~\cite{10878697}.

\subsubsection{Performance and Baseline Comparison}
We evaluate the performance of each method based on 2,000 simulated landing trials in terms of prediction accuracy $e_p = \frac{1}{T} \sum_{t=1}^{T} \| \hat{x}_t - x_t \|_2$, landing deviation $\Delta p = \| P_{\text{land}} - P_{\text{target}} \|_2$, success rate, and computational latency. The success criteria were defined as achieving a landing positional deviation of less than\,\si{5}{cm} and an attitude error of less than $10^\circ$. The computational efficiency was assessed based on the total system latency when executed on an embedded platform, i.e., NVIDIA Jetson Orin NX. 

As shown in Table~\ref{tab:benchmark}, SpecFuse demonstrates superior performance under high-fidelity marine disturbances. By coupling spectral analysis with learning-augmented control, the system achieves a prediction error ($e_p$) of 3.2 cm and a landing deviation ($\Delta p$) of 4.8 cm. This represents a reduction in error of 44\% and 48\%, respectively, compared to the second-best baseline, FFT-KF. While NMPC and MPC-VIO struggle with coupled, non-periodic disturbances, our dual-domain architecture maintains a 98.8\% success rate and the lowest computational latency (82 ms), validating its readiness for real-time embedded deployment.

Table~\ref{tab:computation} provides a granular breakdown of the computational footprint on the NVIDIA Jetson Orin NX. Although the sampling-based HPO-RRT* planner is the primary resource consumer, the Learning-Based Velocity Control remains highly efficient (6.1 ms), facilitating rapid control loop updates during the critical final descent. With a total system latency of 81.9 ms and a modest 268 MB peak RAM footprint, the framework stays well within embedded hardware constraints. This enables a response rate significantly faster than the dominant wave period, ensuring stability. Qualitatively, Fig.~\ref{fig:replanning} illustrates the adaptive planner dynamically adjusting to wave-induced motions to maintain a safe landing trajectory.

\begin{table}[t]
\caption{Computational Efficiency Analysis on Embedded Hardware}
\label{tab:computation}
\centering
\begin{tabular}{lcc}
\toprule
\textbf{Module} & \textbf{Latency (\si{ms})} & \textbf{Peak RAM (\si{MB})} \\
\midrule
{Spectral} Prediction & 34.2 & 82 \\
HPO-RRT* Planning~\cite{9512957} & 41.7 & 157 \\
{Learning-Based} Velocity Control & 6.1 & 29 \\
\midrule
Total System & 81.9 & 268 \\
\bottomrule
\end{tabular}
% \vspace{-0.2cm}
\end{table}

\begin{figure}[t]
\centering
\includegraphics[width=0.85\linewidth]{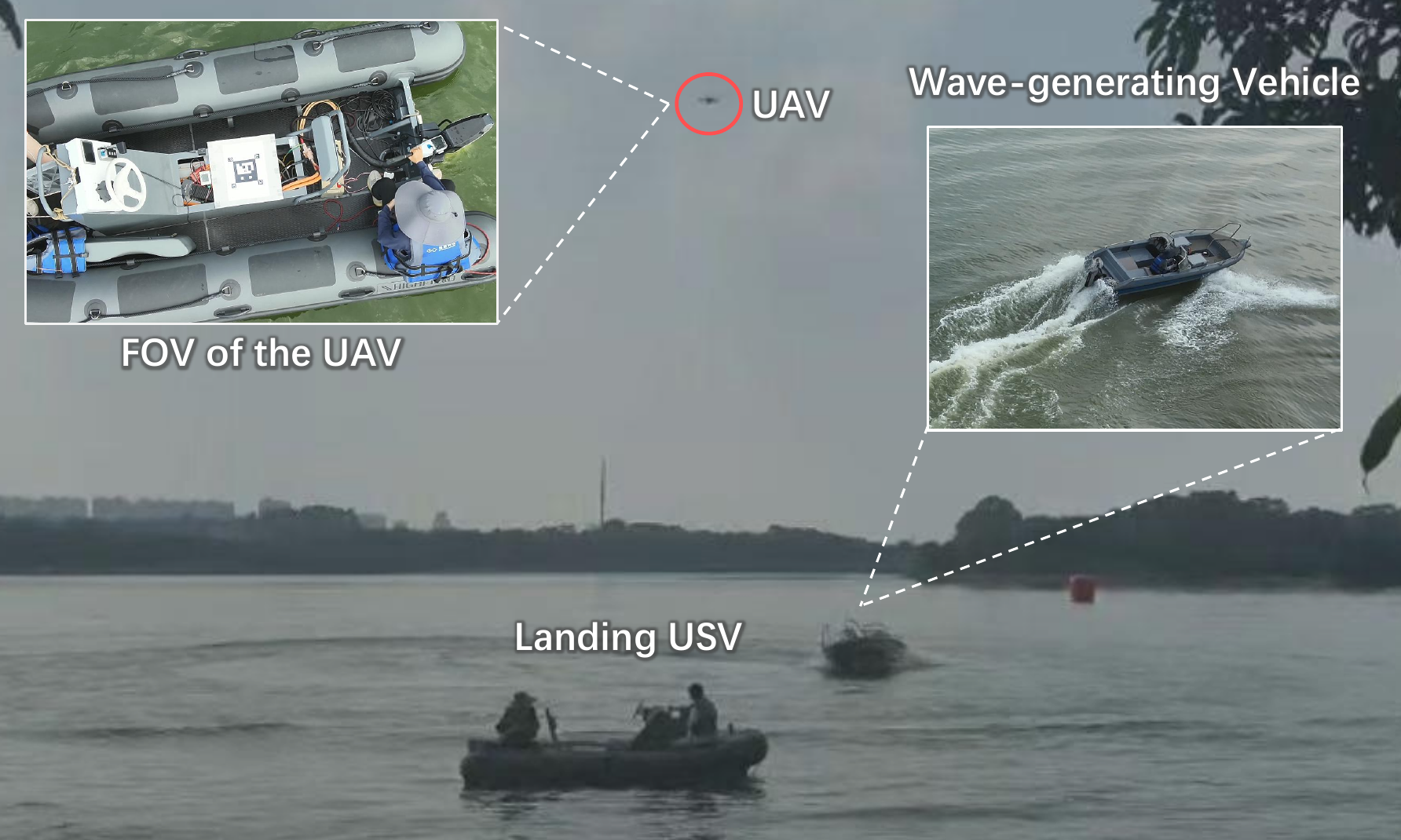}
\caption{Experimental setup in the outdoor environment. A wave-generating vehicle follows a predefined trajectory around the USV, with the UAV's field of view indicated.}
\label{fig:real_world_env}
\vspace{-0.3cm}
\end{figure}

Despite the system's high reliability, a forensic analysis of the 27 failure instances (1.3\% of the total trials) reveals distinct operational boundaries. The failures were predominantly categorized into two modes: \textbf{Wind shear transients} (gusts $>15\,\text{m/s}$), which accounted for 63\% of failures, and \textbf{High-frequency resonance} ($>3\,\text{Hz}$ oscillations), accounting for 28\%. The susceptibility to extreme wind shear suggests occasional excursions beyond the UAV's thrust saturation limits, while the resonance failures indicate rare scenarios where the wave frequency exceeded the effective bandwidth of the {spectral} prediction module. These findings informed the safety protocols for real-world validation, establishing strict cut-off thresholds for environmental conditions during field experiments.

\subsection{Real-World Experimental Results}
\subsubsection{Experimental Setup}
To quantify the sim-to-real gap and validate system robustness, experiments were conducted using a custom \SI{40}{cm} quadrotor platform equipped with a Livox Mid-360 LiDAR and an NVIDIA Jetson Orin NX. The validation strategy focused on unstructured field trials where a 5.5-m speedboat was deployed to generate non-periodic, high-energy wakes, illustrated in Fig.~\ref{fig:real_world_env}. 
The environmental parameters, wind speed and wave height, encountered during the outdoor lake trials were estimated in real time from the onboard IMUs of the UAV and the USV, respectively (see Table \ref{tab:outdoor_lake_env_params}). This highly dynamic, non-periodic environment rigorously tested the framework's performance against complex, multi-source wave spectra, pushing the operational envelope near the identified simulation failure thresholds.

\subsubsection{Outdoor Lake Experiments}
\begin{figure}[t]
    \centering
    \subfigure[Third-person view]{
        \includegraphics[width=0.3867\linewidth]{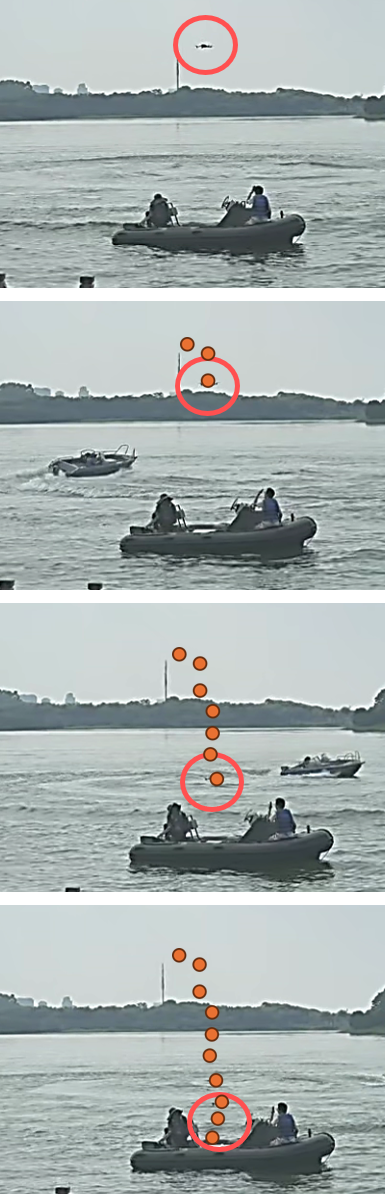}
        \label{fig:tpv_landing}
    }
    \hfill
    \subfigure[First-person view]{
        \includegraphics[width=0.52\linewidth]{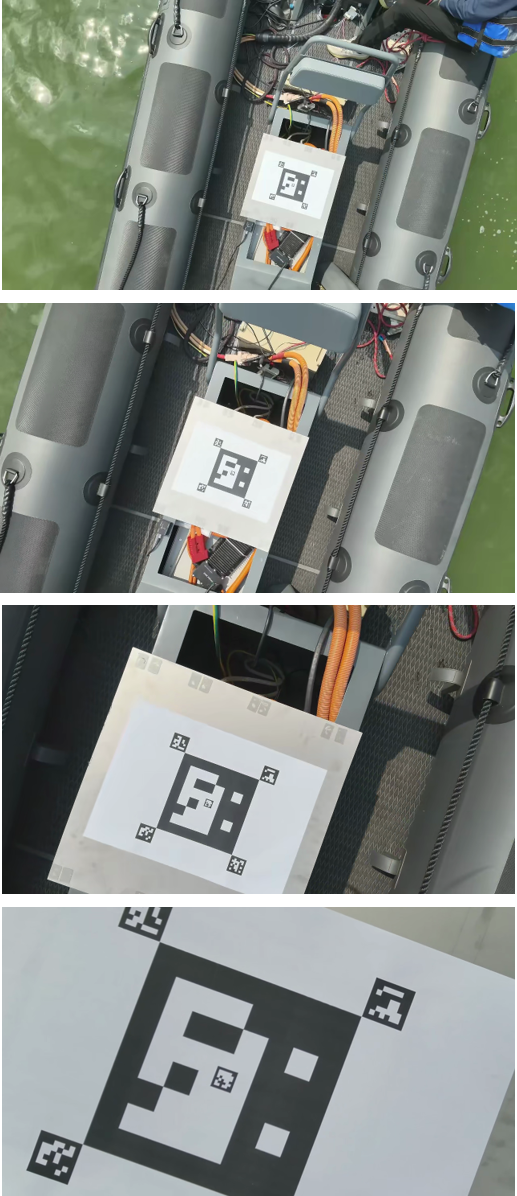}
        \label{fig:fpv_landing}
    }
    \caption{{Autonomous UAV landing on an oscillating marine platform in an outdoor lake environment. The left column displays the third-person view with the UAV highlighted by red circles, while the right column shows the corresponding onboard downward-facing camera views of the landing target.}}
    \label{fig:outdoor_landing_visuals}
\vspace{-0.5cm}
\end{figure}

\begin{table}[t]
\caption{Calculated Environmental Parameters for Outdoor Lake Experiments}
\label{tab:outdoor_lake_env_params}
\centering
\setlength{\tabcolsep}{4pt} % Retaining for compactness
\begin{tabular}{cccc|cccc} % Vertical line added here (|)
\toprule
\textbf{No.} & \textbf{\shortstack{Mean\\Wind}} & \textbf{\shortstack{Max\\Wind}} & \textbf{\shortstack{Wave\\Height}} &
\textbf{No.} & \textbf{\shortstack{Mean\\Wind}} & \textbf{\shortstack{Max\\Wind}} & \textbf{\shortstack{Wave\\Height}} \\
& {(\si{m/s})} & {(\si{m/s})} & {(\si{m})} &
& {(\si{m/s})} & {(\si{m/s})} & {(\si{m})} \\
\midrule
1 & 7.2 & 9.4 & 1.5 & 5 & 7.5 & 9.2 & 1.6 \\
2 & 8.1 & 10.2 & 1.7 & 6 & 10.5 & 12.7 & 2.0 \\
3 & 6.8 & 8.9 & 1.4 & 7 & 8.7 & 10.9 & 1.8 \\
4 & 9.3 & 11.8 & 1.9 & 8 & 13.2 & 15.6 & 2.3 \\
\bottomrule
\end{tabular}
\vspace{-0.3cm}
\end{table}

As summarized in Table~\ref{tab:outdoor_lake_landing_results}, the system achieved an 87.5\% success rate (7 of 8 trials) in the unstructured lake environment. The single failed trial is further investigated in Section~\ref{failureAnalysis}. Despite the introduction of non-periodic boat wakes and natural wind gusts, the framework maintained a low mean landing deviation of \SI{4.46}{cm}. This performance underscores the robustness of the {spectral-temporal} pipeline against unmodeled disturbances. Notably, in Trial 3, the prediction module sustained high accuracy, 91.0\% in roll and 90.3\% in pitch, demonstrating effective generalization.
A representative landing trial is illustrated in Fig.~\ref{fig:outdoor_landing_visuals}. The third-person perspective in Fig.~\ref{fig:tpv_landing} depicts the UAV’s stable approach and final descent onto the moving USV while navigating wake-induced disturbances from a secondary vessel. Simultaneously, the onboard camera view in Fig.~\ref{fig:fpv_landing} validates the perception system's resilience; it maintains continuous tracking of the deck-mounted AprilTag fiducial marker throughout the sequence. This ensures high-precision relative state estimation despite significant multi-axis platform oscillations, specifically pitching and rolling motions.
\subsubsection{Performance Limits and Safety Verification}\label{failureAnalysis}
The sole failure (Trial 8) occurred under extreme conditions (15.6 m/s wind, $H_s = 2.3$ m), where instantaneous wave dynamics saturated the UAV’s thrust control authority. This boundary case identifies the operational ceiling of our modeling trade-off between computational efficiency and high-frequency noise rejection. Crucially, as the deck inclination surpassed the $35^\circ$ safety threshold, the fail-safe logic triggered a hover maneuver, validating the system's safety-critical robustness.

\begin{table}[t]
\centering
\begin{threeparttable}
\caption{Real-world Landing Results in Outdoor Lake Experiments}
\label{tab:outdoor_lake_landing_results}
\begin{tabular}{lcccc}
\toprule
\textbf{No.} & \textbf{Pred. Err.} & \textbf{Land. Dev.} & \textbf{Land. Time} & \textbf{Succ.} \\
\midrule
1 & 3.1\,cm & 4.2\,cm & 22.3\,s & \checkmark \\
2 & 3.9\,cm & 5.1\,cm & 23.1\,s & \checkmark \\
3 & 2.8\,cm & 3.8\,cm & 21.9\,s & \checkmark \\
4 & 3.6\,cm & 4.7\,cm & 22.8\,s & \checkmark \\
5 & 3.0\,cm & 4.0\,cm & 22.5\,s & \checkmark \\
6 & 4.2\,cm & 5.4\,cm & 23.4\,s & \checkmark \\
7 & 2.9\,cm & 3.9\,cm & 22.0\,s & \checkmark \\
8\tnote{*} & 3.5\,cm & 4.6\,cm & --- & $\times$ \\
\midrule
\textbf{Mean} & \textbf{3.5\,cm} & \textbf{4.46\,cm} & \textbf{22.59\,s} & \textbf{87.5\%} \\
\bottomrule
\end{tabular}
\begin{tablenotes}
    \small
    \item[*]Detailed failure analysis is provided in Section~\ref{failureAnalysis}.
\end{tablenotes}
\end{threeparttable}
\vspace{-0.5cm}
\end{table}

\section{Conclusion}
This study establishes a robust prediction-driven control paradigm for autonomous UAV landing on {oscillating marine platforms}, effectively bridging the gap between {spectral} motion analysis and real-time trajectory optimization. Extensive maritime validation confirms the system's reliability, demonstrating over 90\% prediction accuracy and sub-100 ms latency even under stochastic wave disturbances. Beyond the immediate landing task, this research advances safety assurance methodologies for coupled dynamic systems by explicitly defining operational boundaries within the frequency domain. Future work will prioritize enhancing scalability and environmental adaptability; specifically, we aim to implement FPGA-based acceleration and physics-informed learning to tackle Sea State 5 conditions~\cite{bitner2016sea}, while extending the framework to support distributed multi-UAV coordination for complex amphibious missions.

\bibliographystyle{IEEEtran}
\bibliography{references}

@ARTICLE{9998066,
  author={Gupta, Parakh M. and Pairet, Eric and Nascimento, Tiago and Saska, Martin},
  journal={IEEE Robotics and Automation Letters}, 
  title={Landing a {UAV} in Harsh Winds and Turbulent Open Waters}, 
  year={2023},
  volume={8},
  number={2},
  pages={744-751},
  doi={10.1109/LRA.2022.3231831}}

@ARTICLE{10045543,
  author={Mao, Jeffrey and Nogar, Stephen and Kroninger, Christopher M. and Loianno, Giuseppe},
  journal={IEEE Transactions on Robotics}, 
  title={Robust Active Visual Perching With Quadrotors on Inclined Surfaces}, 
  year={2023},
  volume={39},
  number={3},
  pages={1836-1852},
  doi={10.1109/TRO.2023.3238911}}

@INPROCEEDINGS{9981489,
  author={Ji, Jialin and Yang, Tiankai and Xu, Chao and Gao, Fei},
  booktitle={2022 IEEE/RSJ International Conference on Intelligent Robots and Systems (IROS)}, 
  title={Real-Time Trajectory Planning for Aerial Perching}, 
  year={2022},
  volume={},
  number={},
  pages={10516-10522},
  doi={10.1109/IROS47612.2022.9981489}}

@INPROCEEDINGS{8741624,
  author={Kuntz Rangel, Rodrigo and Freitas, Joacy L. and Antônio Rodrigues, Vilmar},
  booktitle={2019 IEEE Aerospace Conference}, 
  title={Development of a Multipurpose Hydro Environmental Tool using Swarms, {UAV} and {USV}}, 
  year={2019},
  volume={},
  number={},
  pages={1-15},
  doi={10.1109/AERO.2019.8741624}}

@INPROCEEDINGS{8206510,
  author={Xiao, Xuesu and Dufek, Jan and Woodbury, Tim and Murphy, Robin},
  booktitle={2017 IEEE/RSJ International Conference on Intelligent Robots and Systems (IROS)}, 
  title={{{UAV}}-assisted {{USV}} visual navigation for marine mass casualty incident response}, 
  year={2017},
  volume={},
  number={},
  pages={6105-6110},
  doi={10.1109/IROS.2017.8206510}}

@ARTICLE{8903530,
  author={Chen, Junting and Gesbert, David},
  journal={IEEE Transactions on Wireless Communications}, 
  title={Efficient Local Map Search Algorithms for the Placement of Flying Relays}, 
  year={2020},
  volume={19},
  number={2},
  pages={1305-1319},
  doi={10.1109/TWC.2019.2952612}}

@ARTICLE{9756345,
  author={Cao, Jiahui and Yang, Zhibo and Tian, Shaohua and Li, Haoqi and Jin, Ruochen and Yan, Ruqiang and Chen, Xuefeng},
  journal={IEEE Transactions on Industrial Electronics}, 
  title={Biprobes Blade Tip Timing Method for Frequency Identification Based on Active Aliasing Time-Delay Estimation and Dealiasing}, 
  year={2023},
  volume={70},
  number={2},
  pages={1939-1948},
  doi={10.1109/TIE.2022.3165252}}

@ARTICLE{7470933,
  author={Zeng, Yong and Zhang, Rui and Lim, Teng Joon},
  journal={IEEE Communications Magazine}, 
  title={Wireless communications with unmanned aerial vehicles: opportunities and challenges}, 
  year={2016},
  volume={54},
  number={5},
  pages={36-42},
  doi={10.1109/MCOM.2016.7470933}}

@ARTICLE{10684104,
  author={Xu, Ruoyu and Jiang, Zixing and Liu, Beibei and Wang, Yuquan and Qian, Huihuan},
  journal={IEEE Transactions on Robotics}, 
  title={Confidence-Aware Object Capture for a Manipulator Subject to Floating-Base Disturbances}, 
  year={2024},
  volume={40},
  number={},
  pages={4396-4413},
  doi={10.1109/TRO.2024.3463476}}

@ARTICLE{10596075,
  author={Wang, Chang and Wang, Jiaqing and Ma, Zhaowei and Xu, Mingjin and Qi, Kailei and Ji, Ze and Wei, Changyun},
  journal={IEEE Transactions on Vehicular Technology}, 
  title={Integrated Learning-Based Framework for Autonomous Quadrotor {UAV} Landing on a Collaborative Moving UGV}, 
  year={2024},
  volume={73},
  number={11},
  pages={16092-16107},
  doi={10.1109/TVT.2024.3425755}}

@INPROCEEDINGS{8901098,
  author={Erunsal, I. Kagan and Martinoli, Alcherio and Ventura, Rodrigo},
  booktitle={2019 International Symposium on Multi-Robot and Multi-Agent Systems (MRS)}, 
  title={Decentralized Nonlinear Model Predictive Control for 3D Formation of Multirotor Micro Aerial Vehicles with Relative Sensing and Estimation}, 
  year={2019},
  volume={},
  number={},
  pages={176-178},
  doi={10.1109/MRS.2019.8901098}}

@ARTICLE{8421746,
  author={Qin, Tong and Li, Peiliang and Shen, Shaojie},
  journal={IEEE Transactions on Robotics}, 
  title={VINS-Mono: A Robust and Versatile Monocular Visual-Inertial State Estimator}, 
  year={2018},
  volume={34},
  number={4},
  pages={1004-1020},
  doi={10.1109/TRO.2018.2853729}}

@INPROCEEDINGS{10878697,
  author={Zhen, XiangYu and Deng, ZhongLiang and Lou, BoYang and Hou, LiuBo and Wei, LiCheng},
  booktitle={2024 11th International Forum on Electrical Engineering and Automation (IFEEA)}, 
  title={Autonomous Navigation Algorithm of Monocular {UAV} Based on Depth Estimation and Robust VIO}, 
  year={2024},
  volume={},
  number={},
  pages={1167-1172},
  doi={10.1109/IFEEA64237.2024.10878697}}

@ARTICLE{8013706,
  author={Chiang, Kai-Wei and Tsai, Guang-Je and Li, Yu-Hua and El-Sheimy, Naser},
  journal={IEEE Geoscience and Remote Sensing Letters}, 
  title={Development of LiDAR-Based {UAV} System for Environment Reconstruction}, 
  year={2017},
  volume={14},
  number={10},
  pages={1790-1794},
  doi={10.1109/LGRS.2017.2736013}}

@INPROCEEDINGS{8484495,
  author={Zhang, MingXi and Li, Qi and Meng, XiangDong and He, YuQing and Luo, HaiTao},
  booktitle={2018 IEEE International Conference on Mechatronics and Automation (ICMA)}, 
  title={Wave Compensator Design Based on Adaptive {FFT} Prediction Algorithm and H$\infty{}$ filtering}, 
  year={2018},
  volume={},
  number={},
  pages={45-50},
  doi={10.1109/ICMA.2018.8484495}}

@INPROCEEDINGS{6950920,
  author={Malkauthekar, M. D.},
  booktitle={Third International Conference on Computational Intelligence and Information Technology (CIIT)}, 
  title={Analysis of Euclidean Distance and Manhattan Distance measure in face recognition}, 
  year={2013},
  volume={},
  number={},
  pages={503-507},
  doi={10.1049/cp.2013.2636}}

@INPROCEEDINGS{10318328,
  author={Gao, Yang and Deng, Xiangyang and Fang, Wei},
  booktitle={2023 IEEE International Conference on Unmanned Systems (ICUS)}, 
  title={Dynamic Modeling and Flight Performance Analysis of Quadrotor {UAV} Under Wind Field Disturbances}, 
  year={2023},
  volume={},
  number={},
  pages={240-245},
  doi={10.1109/ICUS58632.2023.10318328}}

@INPROCEEDINGS{10874309,
  author={Ying, Zheng and Lizhen, Zhang and Jiajia, Yu and Jiafen, Wu},
  booktitle={2024 4th International Signal Processing, Communications and Engineering Management Conference (ISPCEM)}, 
  title={Analysis of Geometric Parameters Error and Simulation of Six-Degree-of-Freedom Industrial Robot}, 
  year={2024},
  volume={},
  number={},
  pages={23-26},
  doi={10.1109/ISPCEM64498.2024.00010}}

@INPROCEEDINGS{9842272,
  author={Poudel, Bidur and Aslami, Pooja and Aryal, Tara and Bhujel, Niranjan and Rai, Astha and Rauniyar, Manisha and Rekabdarkolaee, Hossein Moradi and Tamrakar, Ujjwol and Hansen, Timothy M. and Tonkoski, Reinaldo},
  booktitle={2022 International Symposium on Power Electronics, Electrical Drives, Automation and Motion (SPEEDAM)}, 
  title={Comparative Analysis of State and Parameter Estimation Techniques for Power System Frequency Dynamics}, 
  year={2022},
  volume={},
  number={},
  pages={754-761},
  doi={10.1109/SPEEDAM53979.2022.9842272}}

@ARTICLE{10321664,
  author={Anand, Pritam and Jain, Shantanu and Mishra, Rahul},
  journal={IEEE Sensors Letters}, 
  title={Huber SVR-Based Hybrid Models for Significant Wave Height Forecasting Using Buoy Sensors}, 
  year={2023},
  volume={7},
  number={12},
  pages={1-4},
  doi={10.1109/LSENS.2023.3333967}}

@INPROCEEDINGS{10239764,
  author={Xu, Shibo and Qi, Juntong and Wang, Mingming and Wu, Chong},
  booktitle={2023 42nd Chinese Control Conference (CCC)}, 
  title={Perception-Aware Trajectory Planning of Quadrotor in SE(3)}, 
  year={2023},
  volume={},
  number={},
  pages={4542-4548},
  doi={10.23919/CCC58697.2023.10239764},
  ISSN={1934-1768},
  month={July},}

@INPROCEEDINGS{9512957,
  author={Chen, Jianqing and Yu, Jiyan},
  booktitle={2021 4th International Conference on Advanced Electronic Materials, Computers and Software Engineering (AEMCSE)}, 
  title={An Improved Path Planning Algorithm for {UAV} Based on {RRT}}, 
  year={2021},
  volume={},
  number={},
  pages={895-898},
  doi={10.1109/AEMCSE51986.2021.00182}}

@article{rueda2020selection,
  title={Selection of JONSWAP spectra parameters during water-depth and sea-state transitions},
  author={Rueda-Bayona, Juan Gabriel and Guzm{\'a}n, Andr{\'e}s and Cabello Eras, Juan Jos{\'e}},
  journal={Journal of Waterway, Port, Coastal, and Ocean Engineering},
  volume={146},
  number={6},
  pages={04020038},
  year={2020},
  publisher={American Society of Civil Engineers}
}

@article{ko2025wind,
  title={Wind effects on urban air mobility noise during landing procedures},
  author={Ko, Jeongwoo and German, Brian J and Rauleder, Juergen},
  journal={Aerospace Science and Technology},
  pages={110878},
  year={2025},
  publisher={Elsevier}
}

@article{liu2025udmc,
  author={Liu, Haichao and Chen, Kai and Li, Yulin and Huang, Zhenmin and Liu, Ming and Ma, Jun},
  journal={IEEE Transactions on Intelligent Transportation Systems}, 
  title={{UDMC}: Unified Decision-Making and Control Framework for Urban Autonomous Driving With Motion Prediction of Traffic Participants}, 
  year={2025},
  volume={26},
  number={5},
  pages={5856-5871},
  doi={10.1109/TITS.2025.3551617}
}

@article{gupta2025curvitrack,
  title={CurviTrack: Curvilinear Trajectory Tracking for High-Speed Chase of a {USV}},
  author={Gupta, Parakh M and Proch{\'a}zka, Ond{\v{r}}ej and Nascimento, Tiago and Saska, Martin},
  journal={IEEE Robotics and Automation Letters},
  volume={10},
  number={4},
  pages={3932--3939},
  year={2025},
  publisher={IEEE}
}

@article{li2023integrating,
  title={Integrating dynamic event-triggered and sensor-tolerant control: Application to {USV}-{UAV}s cooperative formation system for maritime parallel search},
  author={Li, Jiqiang and Zhang, Guoqing and Zhang, Xianku and Zhang, Weidong},
  journal={IEEE Transactions on Intelligent Transportation Systems},
  volume={25},
  number={5},
  pages={3986--3998},
  year={2023},
  publisher={IEEE}
}

@article{zhang2025immediate,
  title={An Immediate Update Strategy of Multi-State Constraint Kalman Filter for Visual-Inertial Odometry},
  author={Zhang, Qingchao and Ouyang, Wei and Han, Jiale and Cai, Qi and Zhu, Maoran and Wu, Yuanxin},
  journal={IEEE Robotics and Automation Letters},
  year={2025},
  publisher={IEEE}
}

@article{zheng2025adaptive,
  title={Adaptive Trajectory Learning With Obstacle Awareness for Motion Planning},
  author={Zheng, Huaihang and Tan, Zimeng and Wang, Junzheng and Tavakoli, Mahdi},
  journal={IEEE Robotics and Automation Letters},
  year={2025},
  publisher={IEEE}
}

@article{convens2022safe,
  title={Safe, fast, and efficient distributed receding horizon constrained control of aerial robot swarms},
  author={Convens, Bryan and Merckaert, Kelly and Nicotra, Marco M and Vanderborght, Bram},
  journal={IEEE Robotics and Automation Letters},
  volume={7},
  number={2},
  pages={4173--4180},
  year={2022},
  publisher={IEEE}
}

@article{bitner2016sea,
  title={Sea state conditions for marine structures' analysis and model tests},
  author={Bitner-Gregersen, Elzbieta M and Dong, Sheng and Fu, Thomas and Ma, Ning and Maisondieu, Christophe and Miyake, Ryuji and Rychlik, Igor},
  journal={Ocean Engineering},
  volume={119},
  pages={309--322},
  year={2016},
  publisher={Elsevier}
}

@article{li2025mpc,
  title={MPC-ABCO: An MPC-Based Adaptive Bezier Curve Optimization Framework for UAV-UGV Cooperative Landing},
  author={Li, Haiqi and Qiang, Lei and Wu, Zihao and Chen, Jiajun and Sun, Yinsong and Li, Xinbo},
  journal={IEEE Robotics and Automation Letters},
  year={2025},
  publisher={IEEE}
}
\end{document}